\documentclass{article}

\usepackage{microtype}
\usepackage{graphicx}
\usepackage{subfigure}
\usepackage{booktabs} % for professional tables
\usepackage{mathtools}
\usepackage{epstopdf}
\usepackage{lineno}
\usepackage{authblk}

\usepackage{hyperref}

\usepackage[accepted]{icml2020}

\def\figwidth{7.25cm} %7cm
\def\eqnupspace{-10pt} % -15pt
\def\eqnlowerspace{-5pt} % -8pt
\def\figtocaption{-0.2in} % -0.2in
\def\fighead{-0.0in} % -0.05in
\def\figrear{-0.25in} % -0.3in

\icmltitlerunning{Study on the Large Batch Size Training of Neural Networks Based on the Second Order Gradient}

\begin{document}

\nolinenumbers

\twocolumn[
\icmltitle{Study on the Large Batch Size Training of Neural Networks Based on the Second Order Gradient}




\begin{icmlauthorlist}
\icmlauthor{Fengli Gao}{ime}
\icmlauthor{Huicai Zhong}{ime}
\end{icmlauthorlist}

\icmlaffiliation{ime}{Communication Center, Institute of Microelectronics of the Chinese Academy of Sciences, Beijing, China}
\icmlcorrespondingauthor{Fengli Gao}{gaofengli@ime.ac.cn}

% \author[1]{Alice Smith}
% \author[2]{Bob Jones}
% \affil[1]{Department of Mathematics, University X}

\icmlkeywords{Machine Learning, Large Batch Size, Curvature Radius}

\vskip 0.3in
]



\printAffiliationsAndNotice{}

\begin{abstract}
Large batch size training in deep neural networks (DNNs) possesses a well-known ‘generalization gap’ that remarkably induces generalization performance degradation. However, it remains unclear how varying batch size affects the structure of a NN. Here, we combine theory with experiments to explore the evolution of the basic structural properties, including gradient, parameter update step length, and loss update step length of NNs under varying batch sizes. We provide new guidance to improve generalization, which is further verified by two designed methods involving discarding small-loss samples and scheduling batch size. A curvature-based learning rate (CBLR) algorithm is proposed to better fit the curvature variation, a sensitive factor affecting large batch size training, across layers in a NN. As an approximation of CBLR, the median-curvature LR (MCLR) algorithm is found to gain comparable performance to Layer-wise Adaptive Rate Scaling (LARS) algorithm. Our theoretical results and algorithm offer geometry-based explanations to the existing studies. Furthermore, we demonstrate that the layer wise LR algorithms, for example LARS, can be regarded as special instances of CBLR. Finally, we deduce a theoretical geometric picture of large batch size training, and show that all the network parameters tend to center on their related minima.
\end{abstract}

\section{Introduction}
\begin{figure}[ht]
\vskip \fighead{}
\begin{center}
% \centerline{\includegraphics[width=\columnwidth]{fig_2.eps}}
\centerline{\includegraphics[width=\figwidth]{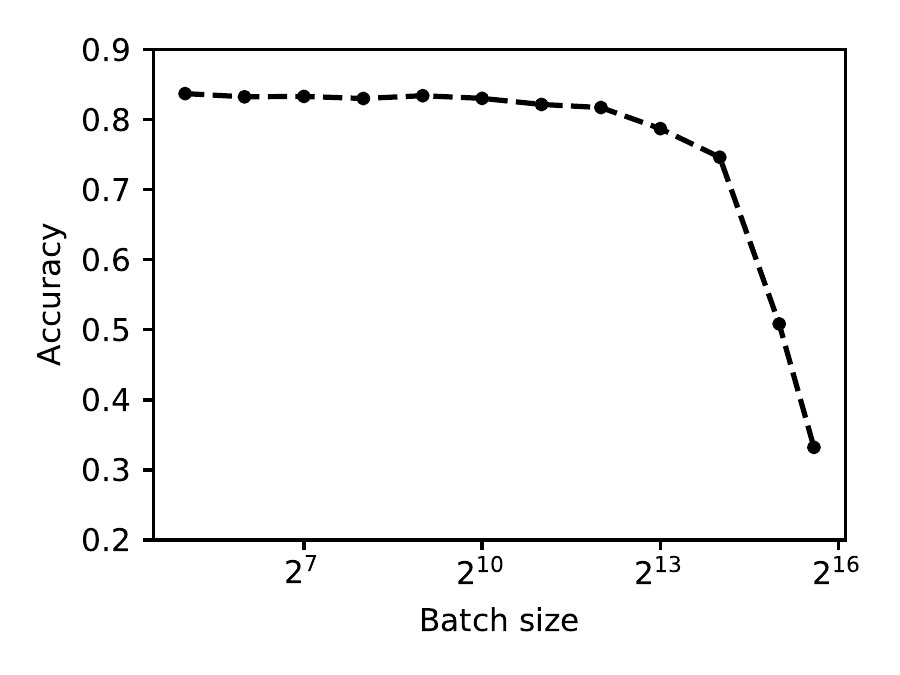}}
\vskip \figtocaption{}
\caption{Impact of batch size on AlexNet accuracy using Cifar10 dataset}
\label{fig_1}
\end{center}
\vskip \figrear{}
\end{figure}

In recent years, there are more and more big data based deep learning tasks, such as ImageNet classification, natural language processing, and face recognition. The datasets often include tens of millions of images, texts, voices or other data, and the related neural networks (NNs) also contain multiple millions of parameters. For such complex tasks, how to improve the training efficiency has attracted mounting attention. It is natural to use multiple Graphics Processing Unit (GPU) as the parallel computing power, which can feed a very large batch size of data into the network at each training iteration. Unfortunately, this large batch size training deteriorates the model performance in respect of accuracy, convergence time, and more. This generalization difficulty has been reported previously \cite{lecun2012efficient,diamos2016persistent,keskar2017on,goyal2017accurate,jastrzebski2018three}, and widely known as a ‘generalization gap’ \cite{keskar2017on}. We reimplement this problem using AlexNet on Cifar10 dataset (Figure \ref{fig_1}).

Many efforts have been devoted to understanding the ‘generalization gap’ \cite{breuel2015the,keuper2016distributed,neyshabur2017exploring,li2017batch,dai2018towards}. Keskar et al. \yrcite{keskar2017on} and Jastrzebski et al. \yrcite{jastrzebski2018finding} found that large batch size training is more likely to converge to sharp minima compared to small batch size training. In this vein, Wen et al. \yrcite{wen2018smoothout:} and Haruki et al. \yrcite{haruki2019gradient} proposed noise injecting methods to eliminate sharp minima. Jastrzebski et al. \yrcite{jastrzebski2018three} reported that higher values of the ratio of learning rate (LR) to batch size can lead to flatter minima.

Hoffer et al. \yrcite{hoffer2017train} put forward a new hypothesis, ‘random walk on a random landscape’, and experimentally demonstrated that the ‘generalization gap’ can be primarily attributed to the reduced number of parameter updates. This hypothesis suggests that longer training time helps to achieve better generalization.

LR scheduling is a class of methods that could alleviate the ‘generalization gap’. Goyal et al. \yrcite{goyal2017accurate} tested a ‘learning rate warm-up’ scheme, in which a small ‘safe’ LR was initialized and linearly scaled to a target ‘base’ LR. Using this method, they successfully compress the training time of ImageNet to less than 1 hour. MegDet, an object detector, obtained the state of the art performance using a combined policy: warmup LR and Cross-GPU batch normalization \cite{peng2017megdet:}. You et al. \yrcite{you2019large-batch} proposed an improved approach of warm-up, linear-epoch gradual-warmup (LEGW), which gains better generalization. Smith \& Topin \yrcite{smith2017super-convergence:} described a phenomenon named ‘super-convergence’ with periodic LR modulation. Li et al. \yrcite{li2019towards} demonstrated that NN models with small LR first generalize to low-noise and hard-to-fit patterns, and Jastrzebski et al. \yrcite{jastrzebski2018three} explained the ratio of LR to batch size is a key determinant of statistical gradient descent (SGD) dynamics.

Batch size scheduling is regarded as the counterpart of LR scheduling. The largest useful batch size can be predicted by the gradient noise scale \cite{mccandlish2018an}. Increasing batch size helps find flatter minima \cite{devarakonda2017adabatch:,smith2018don't}. Mikami \& Suganuma \yrcite{Mikami2018imageNet} have trained ImageNet in 224 seconds using a predetermined batch size scheduling.

The discovery of the unbalance distribution of both parameters and gradients across layers in a network is an important breakthrough for large batch size training. The generalization is markedly enhanced in case of setting a layer-wise LR, which is determined by some form of statistical values of the inlayer parameters and gradients. You et al. \yrcite{you2017large} firstly built an algorithm, ‘Layer-wise Adaptive Rate Scaling (LARS)’, to extend ImageNet training batch size to 32 K. In their sequential work, they showed that using LARS, the ImageNet can be even trained in a few minutes \cite{you2017scaling,you2018imagenet}. Two relatives of LARS, ‘PercentDelta’ and ‘LAMB’, have achieved better generalization \cite{abuelhaija2017proportionate,you2019reducing}. Using ‘LAMB’ optimizer, You et al. \yrcite{you2019reducing} have reduced the pre-training time of BERT to 76 minutes.

In parallel with the above-mentioned approach, new methods such as adding noise to gradients \cite{wen2019interplay}, applying second-order information \cite{martens2015optimizing,yao2018large,adya2018nonlinear}, using local SGD instead of large batch size \cite{lin2018don't}, and batch augmentation \cite{hoffer2017train} have also attracted a lot of interest.

In this study, we explore the evolution of the basic geometry properties of NNs with varying batch sizes. Such geometrical properties include parameter, gradient, and loss. Our main work include:

\begin{itemize}
\item First, we make a theoretical and experimental analysis on the geometry limit of large batch size training.
\item Based on the work above, the authors propose a guidance to improve ‘generalization gap’.
\item Then we design a parameter update algorithm based on the approximated loss curve curvature.
\item In the last part, we provide theoretical backbones for the major part of the existing research results of ‘generalization gap’, and investigate the geometry picture of large batch size training.
\end{itemize}

% \end{Introduction}
\section{Theoretical and Experimental Analysis}

Figure \ref{fig_1} raises a highly concerned problem that large batch size deteriorates the training performance. To avoid subjective bias and search for an in-depth insight, we focus on inspecting of some fundamental network properties, including gradient, parameter update step length and loss update step length with varying batch sizes. If there’s no special specifications, all the experiments in this paper are carried out on Pytorch, using AlexNet and Cifar10 dataset, with SGD as the default optimizer.

\subsection{Gradient}
\begin{figure}[ht]
\vskip \fighead{}
\begin{center}
\centerline{\includegraphics[width=\figwidth]{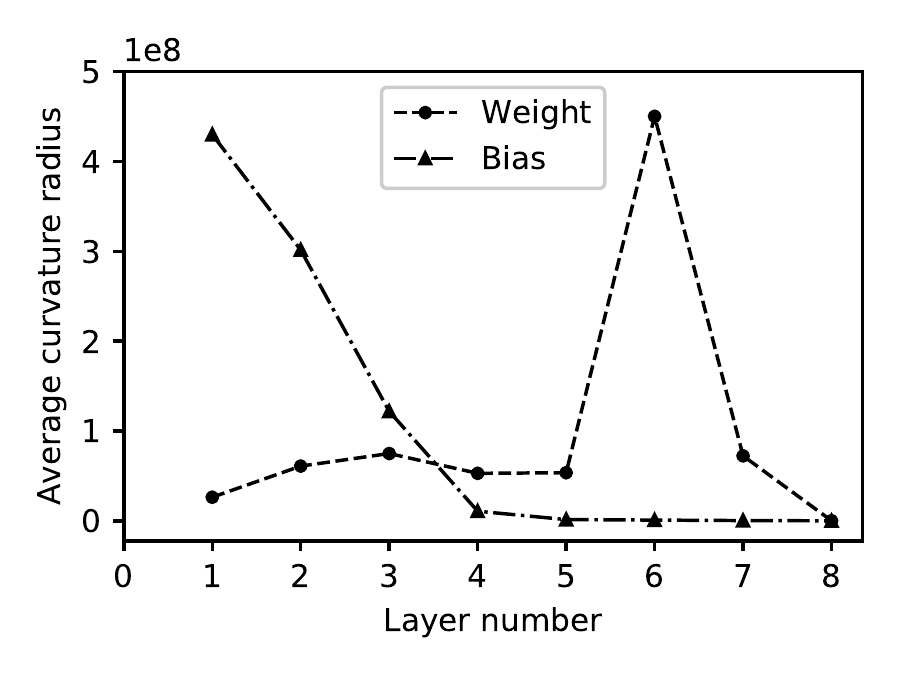}}
\vskip \figtocaption{}
\caption{Curvature variations across the layers of AlexNet. The average curvature radius of Bias is artificially enlarged by10 times.}
\label{fig_2}
\end{center}
\vskip \figrear{}
\end{figure}

We probe the impact of batch size on gradient by combining theory and experiments. The curvature of loss determines the parameter update amplitude in terms of LR, and in a update step, the related parameter will inch closer to a nearby minima of the loss curve, with a update step length given by the product of gradient and LR. We find that curvature across layers in a network varies remarkably (Figure \ref{fig_2}), and thus it is reasonable to treat gradient in a manner of layer independent. Theoretically, the gradient of each parameter is the average value of the samples in a whole batch size, and we suppose here that the distribution of gradient on a layer is Gaussian especially in the initial training epochs:

\vskip -15pt %\eqnupspace
\begin{equation}
\label{eqn_1}
g^k_i\sim N(\mu,\sigma)
\end{equation}
\vskip \eqnlowerspace
where \(g_i^k\) is the \(i\)th gradient corresponding to the \(k\)th sample in a given layer; \(\mu\) and \(\sigma\) denote the mean and standard deviation of the distribution, respectively. \(\mu=0\) is another assumption considering the parameters evenly distributed near the minima. \(\sigma\) is treated as a constant, which relates to both the average curvature of the layer and the structure of the network. The mean gradient over a batch of sample is

\vskip \eqnupspace
\begin{equation}
\label{eqn_2}
g_i=\frac{1}{n}\sum_{k=0}^n g^k_i\sim N(0,\frac{\sigma}{\sqrt{n}})
\end{equation}
\vskip \eqnlowerspace

where \(n\) is the batch size. The probability density function is

\vskip \eqnupspace
\begin{equation}
\label{eqn_3}
f(g_i)=\frac{\sqrt{n}}{\sigma \sqrt{2\pi}}e^{-\frac{1}{2}(\frac{g_i}{\sigma \sqrt{n}})^2}
\end{equation}
\vskip \eqnlowerspace

The expectation of the absolute value of \(g_i\) on all gradients in a layer is

\vskip \eqnupspace
\begin{equation}
\label{eqn_4}
    E(\left|g_i \right|)=\int_{-\infty}^\infty f(g_i)\left|g_i \right|d g_i=\frac{2\sigma}{\sqrt{\pi}}\frac{1}{\sqrt{n}}
\end{equation}
\vskip \eqnlowerspace

\begin{figure}[ht]
\vskip \fighead{}
\begin{center}
\centerline{\includegraphics[width=\figwidth]{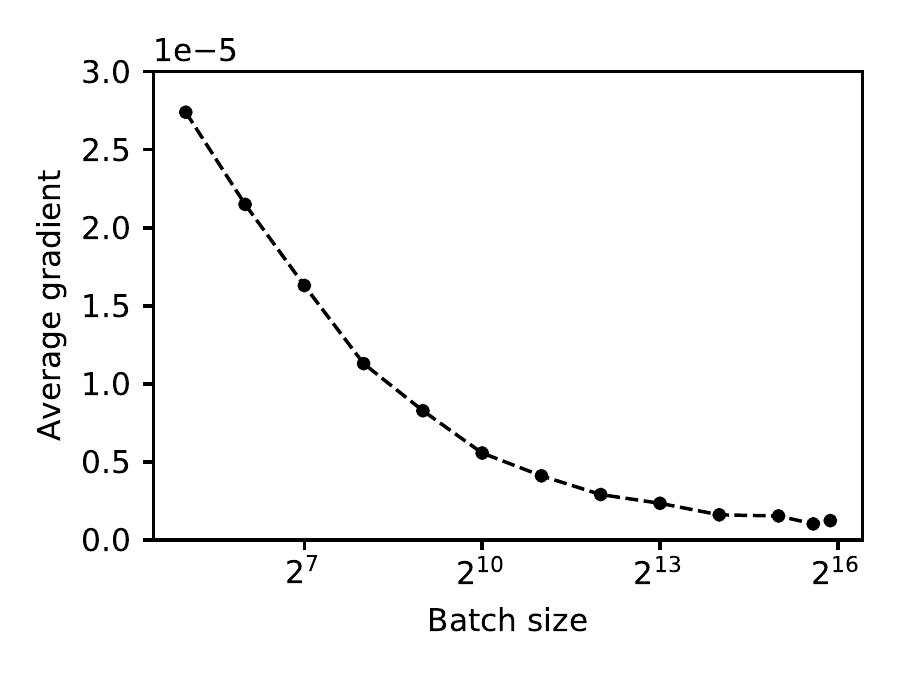}}
\vskip \figtocaption{}
\caption{Impact of batch size on the average gradient. The average gradient is the L1 norm of the gradients divided by the number of the parameters in the first fully connected layer of AlexNet.}
\label{fig_3}
\end{center}
\vskip \figrear{}
\end{figure}

This function predicts that the mean gradient will decrease when batch size increases. To verify this prediction, in the first update step, we compute \(E(\left|g_i \right|)\) of the first fully connected layer of AlexNet on Cifar10 dataset from batch size 32 to 60000, and Figure \ref{fig_3} shows the expected result.

The consistence between theoretical and experimental results indicates that gradients falling to zero with large batch size is immutable, and it is doomed very large batch size training will result in a poor model performance, if there is no adaption on the network architecture.

% \end{Gradient}
\subsection{Parameter}

The gradient vanishing effect of large batch size will definitely lower the parameter update step length. The common parameter update rule is

\vskip -15pt %\eqnupspace
\begin{equation}
    \label{eqn_5}
    \Delta w_i=lr(n) g_i
\end{equation}
\vskip \eqnlowerspace

where \(\Delta w_i\) and \(lr(n)\) are the parameter update step length and the LR with respect to batch size \(n\) , respectively. Combining eqn. \ref{eqn_4}, the average parameter update step length across a specific layer is
\vskip \eqnupspace
\begin{equation}
    \label{eqn_6}
    E(\left| \Delta w_i \right|)=lr(n) E(\left| g_i \right|)=\frac{2\sigma}{\sqrt{\pi}}\frac{lr(n)}{\sqrt{n}}
\end{equation}
\vskip \eqnlowerspace

\begin{figure}[ht]
\vskip \fighead{}
\begin{center}
\centerline{\includegraphics[width=\figwidth]{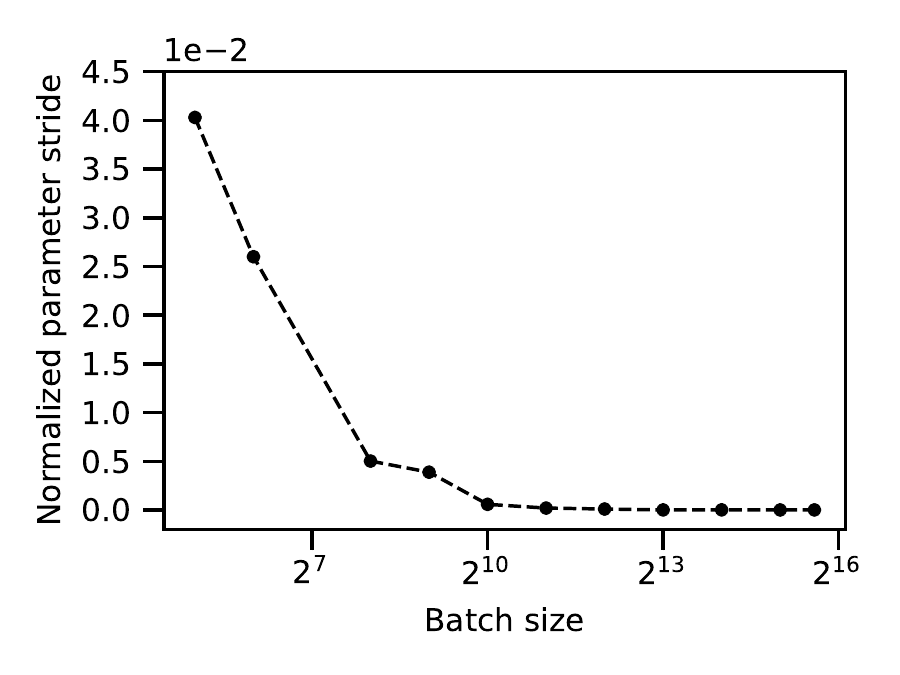}}
\vskip \figtocaption{}
\caption{Impact of batch size on the normalized parameter update step length in the first epoch.}
\label{fig_4}
\end{center}
\vskip \figrear{}
\end{figure}

To validate this finding, we compute the first epoch value of the ‘Normalized parameter stride’, \( E(\left| \Delta w_i \right|) \verb|/| lr(n) \), with varying batch sizes. As shown in Figure \ref{fig_4}, we observe a significant shrink of this value with large batch size.

\begin{figure}[ht]
\vskip \fighead{}
\begin{center}
\centerline{\includegraphics[width=\figwidth]{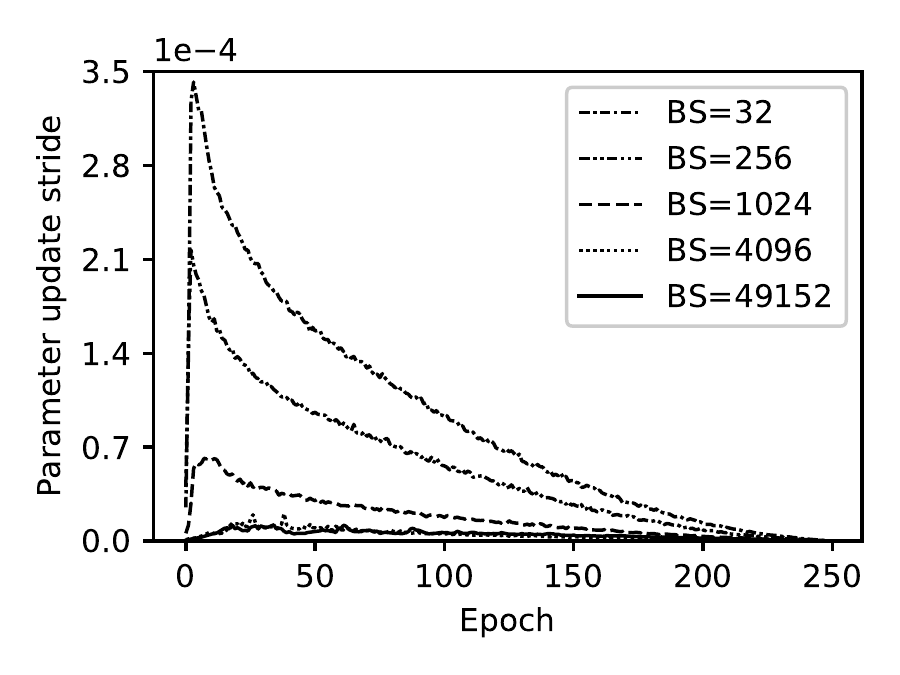}}
\vskip \figtocaption{}
\caption{Parameter update stride from batch-size 32 to batch-size 49152. The area below a curve represents the parameter searching space.}
\label{fig_5}
\end{center}
\vskip \figrear{}
\end{figure}

The parameter searching space accumulated from each update step is a key point for network optimization. It is obvious that the probability to find a better minima is bigger with larger searching space. We record the ‘Parameter update stride’, \( E(\left| \Delta w_i \right|) \), of each epoch during multiple AlexNet training tasks (Cifar10 dataset). As shown in Figure \ref{fig_5}, the area under a single result curve represents the total parameter searching space for a specific batch size. For small batch sizes, the parameter searching space is big enough to obtain an optimal minima for each parameter, while for large batch sizes, it is more likely a small-area minima will be found and thus less likely a good-performance model will be trained.

% \end{Parameter}
\subsection{Loss}

The vanishing of both gradient and parameter update step length will definitely affect the loss update process. We commit a brief theoretical derivation to study the rule change.

\begin{figure}[ht]
\vskip \fighead{}
\begin{center}
% \centerline{\includegraphics[width=\figwidth]{fig_6.pdf}}
\centerline{\includegraphics[width=7cm]{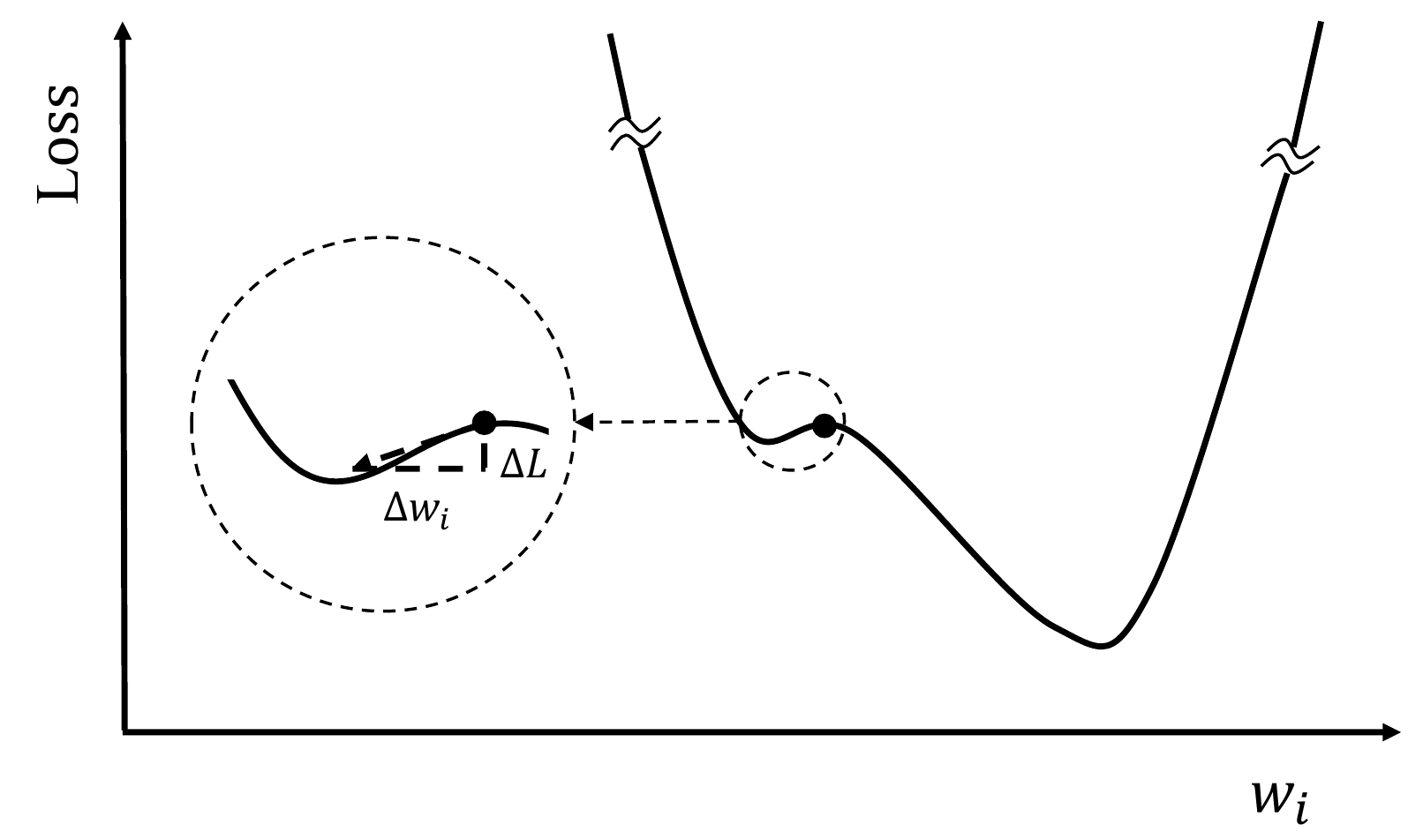}}
% \vskip \figtocaption{}
\vskip -0.1in
\caption{A conceptual sketch of minima, gradient, parameter update step length, and loss update step length.}
\label{fig_6}
\end{center}
\vskip \figrear{}
\end{figure}

Referring the geometry relations of \( \Delta L \) and \( \Delta w_i \) in Figure \ref{fig_6}, the loss decrement induced by a single parameter update step is

\vskip -15pt %\eqnupspace
\begin{equation}
    \label{eqn_7}
    \Delta L_i=\Delta w_i g_i=lr(n) (g_i)^2
\end{equation}
\vskip \eqnlowerspace

Then the average loss update step length across the batch of samples is

\vskip \eqnupspace
\begin{equation}
    \label{eqn_8}
    E(\Delta L_i )=lr(n)E((g_i)^2 )=\sigma^2 \frac{lr(n)}{n}
\end{equation}
\vskip \eqnlowerspace

\begin{figure}[ht]
\vskip \fighead{}
\begin{center}
\centerline{\includegraphics[width=\figwidth]{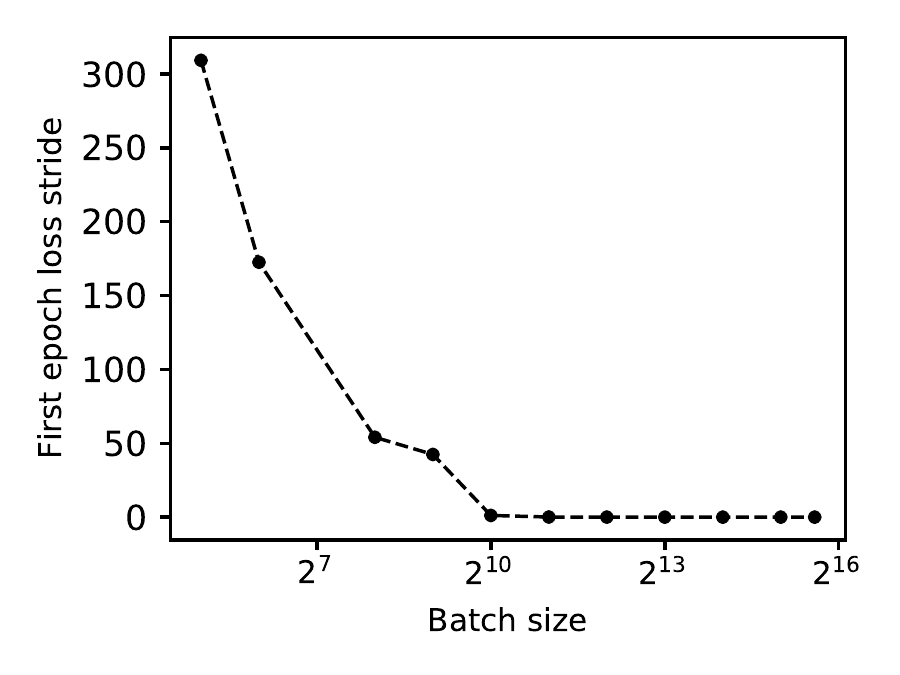}}
\vskip \figtocaption{}
\caption{Impact of batch size on the normalized first epoch loss update length}
\label{fig_7}
\end{center}
\vskip \figrear{}
\end{figure}

This equation gives the theoretical results of the average loss decrement in a single update step. To check the impact of large batch size training on loss changing, we compute the ‘First epoch loss stride’, \( E(\Delta L_i ) / lr(n) \), in the first epoch of training process, and the results shown in Figure \ref{fig_7} are consistent with our theoretical predictions.

\begin{figure}[ht]
\vskip \fighead{}
\begin{center}
\centerline{\includegraphics[width=\figwidth]{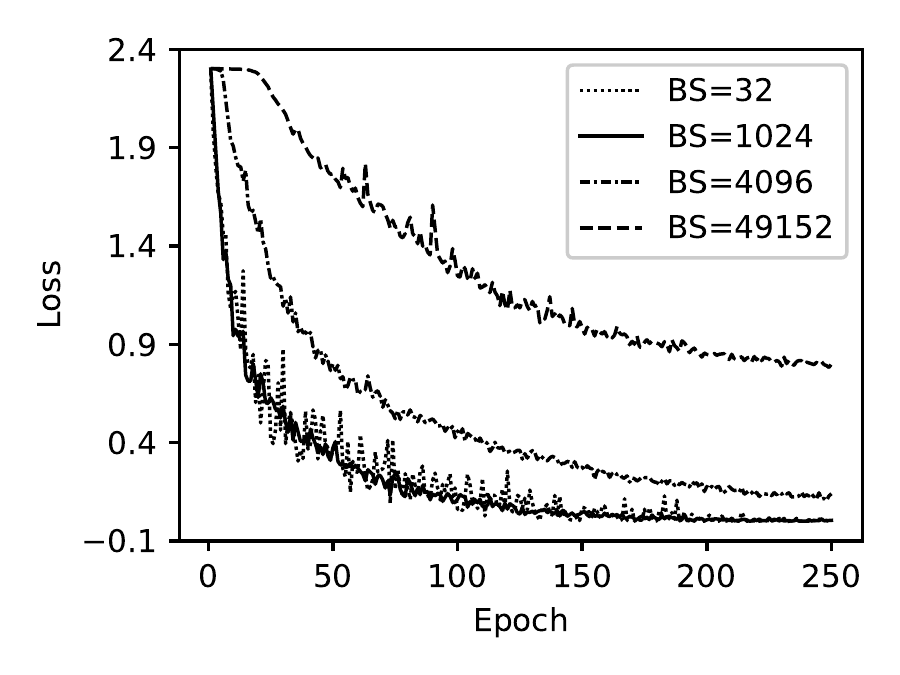}}
\vskip \figtocaption{}
\caption{Loss curves from batch-size 32 to batch-size 49152}
\label{fig_8}
\end{center}
\vskip \figrear{}
\end{figure}

In addition to this experiment, we also compare the whole training process with varying batch sizes in Figure \ref{fig_8}. Besides the loss curves changing as expected, we observe a phenomenon that for both small and large batch sizes, the fluctuation amplitude of the loss curves are big, while for medium batch size, the amplitude are relatively small. We preliminary judge that there are two factors that can lead to network instability: the small parameter update step length in large batch-size training and the big randomness in small batch-size training.

% \end{Loss}
% \end{Theoretical and Experimental Analysis}
\section{Methods to Enlarge the Gradient}

In Section 2, we demonstrate the disadvantages of large batch size training originate from gradient vanishing effect. Thus, on the other hand, it also provides a clue to improve the performance of NN training. We have tried a lot of methods to enlarge the gradient, including discarding small-loss samples, real time tuning of batch size in a training process, discarding small value gradients, setting larger dropout ratio, and selecting sharper loss function. Two methods will be described in the following.

\subsection{Discarding Small-loss Samples}

We conjecture that the input samples with higher losses will produce larger gradients, and we design an experiment to compute the average gradients of the second fully connected layer of AlexNet.

\begin{figure}[ht]
\vskip \fighead{}
\begin{center}
\centerline{\includegraphics[width=\figwidth]{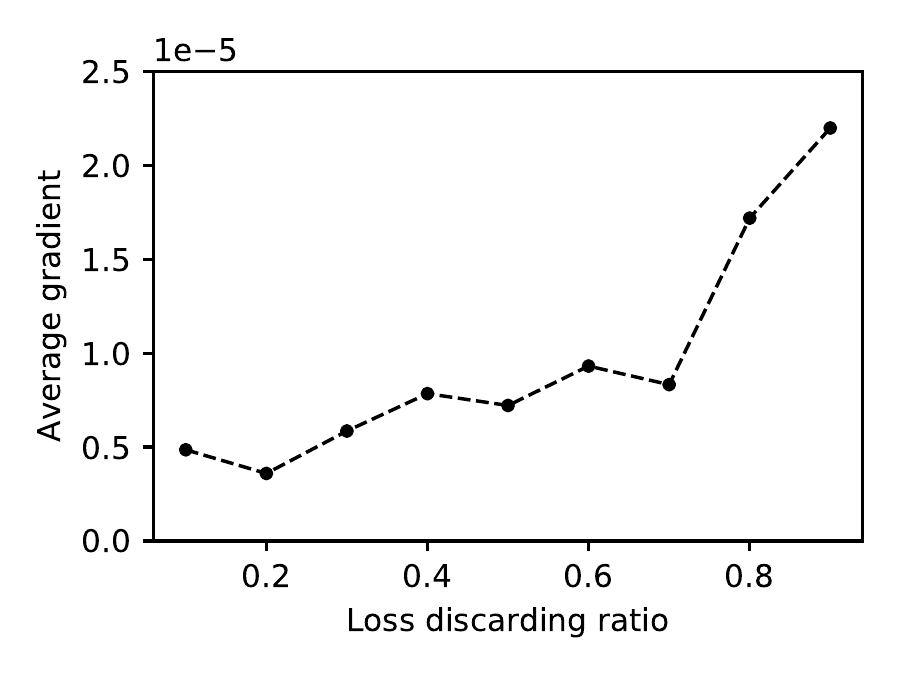}}
\vskip \figtocaption{}
\caption{Impact of loss discarding ration on average gradient. The small-loss samples are discarded in each experiment. The tests are carried out in AlexNet on Cifar10 dataset with batch-size 8192 and update-step 1.}
\label{fig_9}
\end{center}
\vskip \figrear{}
\end{figure}

In the first step, a set of tests are finished, with each one discards p\% of small-loss samples. The loss defined here is computed using a single input sample in a batch size, and the discarding ratio, p\%, changes equally from 10\% to 90\%. Figure \ref{fig_9} shows the average gradients with varying discarding ratios, which illustrates discarding small-loss samples will enlarge the gradient.

The next step is to test the actual model training. We compare the baseline to our model for both batch-size 2000 and 8000. The training parameters in our model are the same as the baseline except discarding 30\% small-loss samples in the first 100 epochs. The result shows an observable improvements for both batch-size 2000 and 8000.

A point to note here is that for batch size larger than 8000, the effect of discarding small-loss samples is not significant, which needs further research.

\begin{figure}[ht]
\vskip \fighead{}
\begin{center}
\centerline{\includegraphics[width=\figwidth]{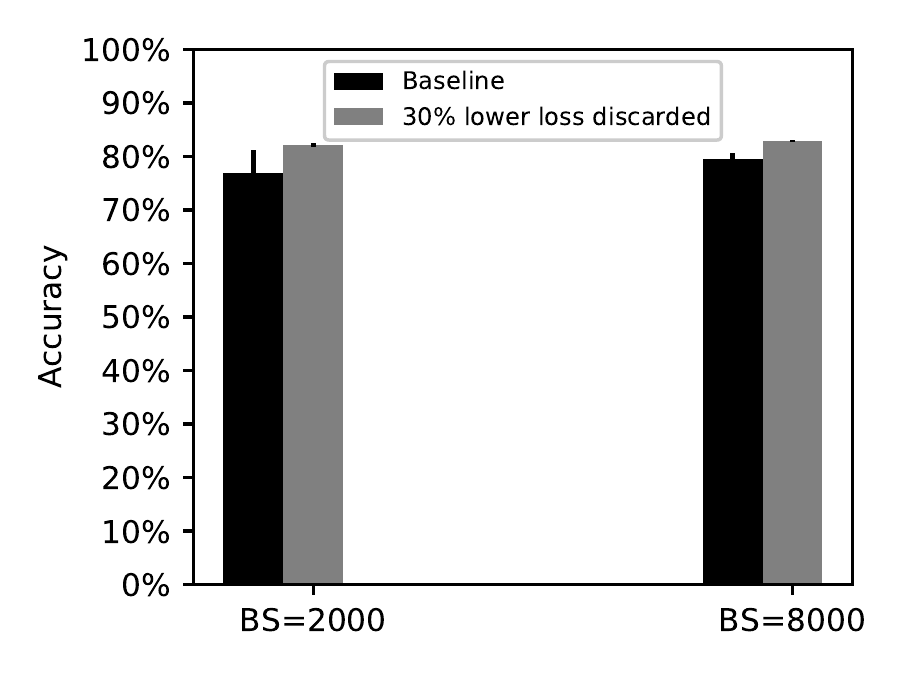}}
\vskip \figtocaption{}
\caption{Training accuracy of baseline and the model with 30\% lower loss discarded. The vertical lines indicate the standard deviation. The baseline model is AlexNet on Cifar10 dataset using SGD optimizer.}
\label{fig_10}
\end{center}
\vskip \figrear{}
\end{figure}

% \end{Discarding Small-loss Samples}
\subsection{Real Time Tuning of Batch Size in a Training Process}

As shown in Figure \ref{fig_3}, a smaller batch size produces larger gradients, so it is worth trying to introduce a different batch size in some steps of a training process. Mikami \& Suganuma \yrcite{Mikami2018imageNet} already used this method to achieve better generalization. In this part, we show some simple comparative experiments as a preliminary exploration.

We use the same training parameters as baseline except in the first epoch, the batch-size and LR are tuned down to 512 and 0.005, respectively, which will be reset to 8192 and 0.05 from the second epoch.

\begin{figure}[ht]
\vskip \fighead{}
\begin{center}
\centerline{\includegraphics[width=\figwidth]{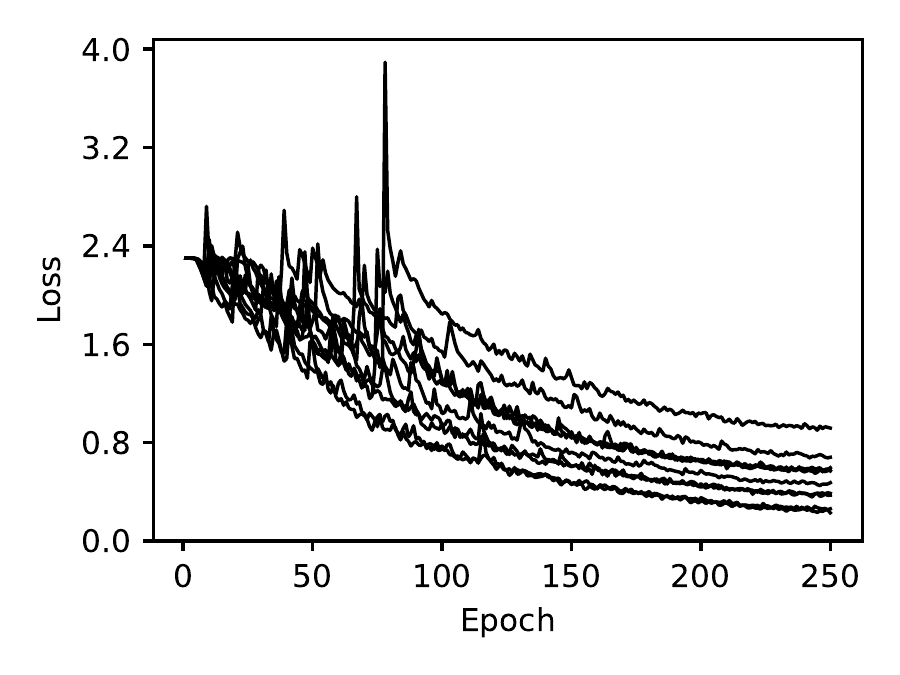}}
\vskip \figtocaption{}
\caption{Multiple loss curves of baseline model. The baseline model is AlexNet on Cifar10 dataset with batch-size 8192 and use SGD as optimizer.}
\label{fig_11}
\end{center}
\vskip \figrear{}
\end{figure}

\begin{figure}[ht]
\vskip \fighead{}
\begin{center}
\centerline{\includegraphics[width=\figwidth]{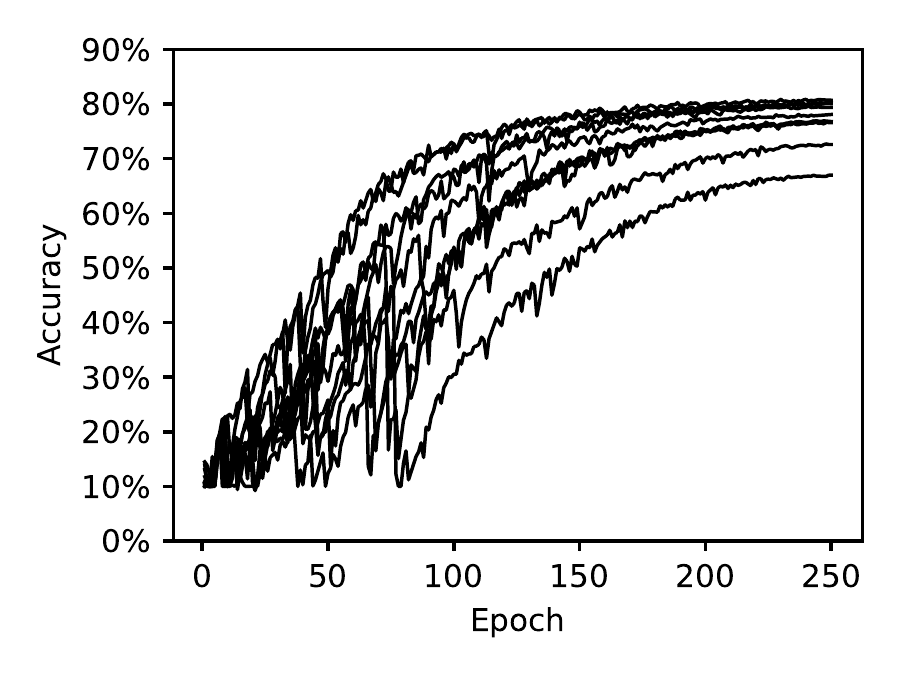}}
\vskip \figtocaption{}
\caption{Multiple accuracy curves of baseline model}
\label{fig_12}
\end{center}
\vskip \figrear{}
\end{figure}

\begin{figure}[ht]
\vskip \fighead{}
\begin{center}
\centerline{\includegraphics[width=\figwidth]{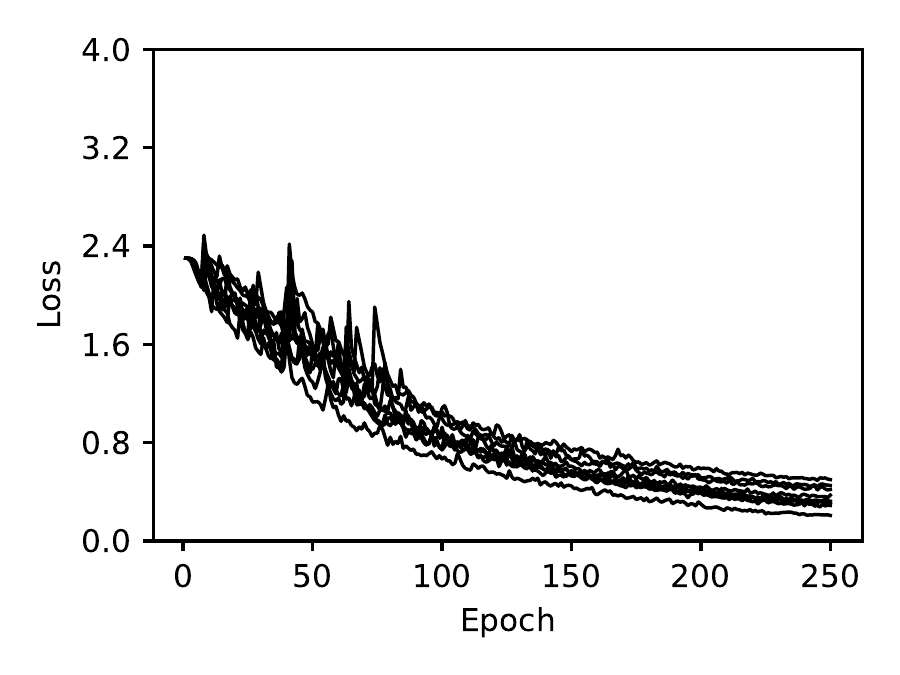}}
\vskip \figtocaption{}
\caption{Multiple loss curves with first-epoch batch size reduced}
\label{fig_13}
\end{center}
\vskip \figrear{}
\end{figure}

\begin{figure}[ht]
\vskip \fighead{}
\begin{center}
\centerline{\includegraphics[width=\figwidth]{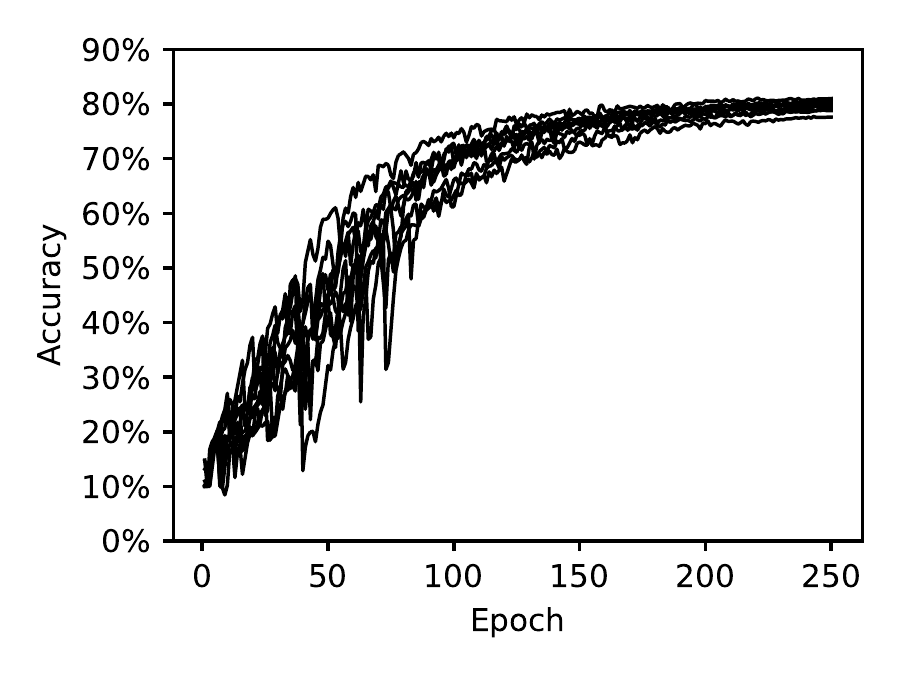}}
\vskip \figtocaption{}
\caption{Multiple accuracy curves with first-epoch batch size reduced}
\label{fig_14}
\end{center}
\vskip \figrear{}
\end{figure}

We have recorded a group of test results for both baseline model and our model (Figure \ref{fig_11}, \ref{fig_12}, \ref{fig_13}, \ref{fig_14}). For the baseline model, the small update step makes the training process more stumble. The big fluctuation in the first 80 epochs of both loss and accuracy curves hinder the subsequent training and results in models with big performance dispersion. The intrinsic reason is related to the small parameter update step length.

While for our model, the training performance is clearly improved, as shown in Figure \ref{fig_13} and \ref{fig_14}. By decreasing the batch size in the first epoch, the update step length is amplified, which enables searching a broader space for a better loss curve minima. 

% \end{Real Time Tuning of Batch Size in a Training Process}
% \end{Methods to Enlarge the Gradient}
\section{Parameter Update Rules Based on Curvature of Loss Curve}

Besides the gradient vanishing effect, the non-uniform distribution of loss curve curvature across layers in a network is another important factor affecting the performance of large batch size training. As shown in Figure \ref{fig_2}, the curvature heterogeneity is significant, and needs pay more attention to. From a geometric point of view, the LR of a specific parameter should be consistent with the loss curve curvature where the parameter located. In case of small batch size, due to the large parameter searching space (Figure \ref{fig_5}), it is applicable to set a global LR, while for large batch size, the decreasing parameter update step length (Figure \ref{fig_4}) makes the training highly sensitive to LR, and thereby, a customized LR setting method is necessary.

\subsection{Curvature-based LR Setting Method and its Approximation}

The best way to update parameters is to set a specific LR for each parameter based on its curvature. The radius of curvature of parameter \( w_i \) is

\vskip \eqnupspace
\begin{equation}
    \label{eqn_9}
    R_i = \left| \frac{(1+(\frac{d L}{d w_i})^2)^{\frac{3}{2}}}{\frac{d^2 L}{d w_i^2}} \right|
\end{equation}
\vskip \eqnlowerspace

where \( R_i \) is the radius of curvature, \( dL/dw_i) \) is the first order gradient and \( dL/dw_i=g_i \), and \( d^2 L/d w_i^2 \) is the second order gradient. Here the LR for \( w_i \) is

\vskip -15pt %\eqnupspace
\begin{equation}
    \label{eqn_10}
    \eta_i=\gamma R_i
\end{equation}
\vskip \eqnlowerspace

where \(\eta_i \) is the LR, and \( \gamma \) is a global hyper parameter. Then follow the rule below to update the parameter,
\vskip \eqnupspace
\begin{equation}
    \label{eqn_11}
    \Delta w_i=\eta_i\frac{d L}{d w_i}
\end{equation}
\vskip \eqnlowerspace

The above steps are the backbone of curvature-based parameter update rule. Though it is a vanilla method, the huge computation work needed to obtain curvature radius in a big NN limits its applications. The main difficulty results from the lack of high efficiency methods to compute second order gradient in almost all deep learning platforms. In this section we develop a method to approximate the curvature radius.

\begin{figure}[ht]
\vskip \fighead{}
\begin{center}
% \centerline{\includegraphics[width=\figwidth]{fig_15.pdf}}
\centerline{\includegraphics[width=6.5cm]{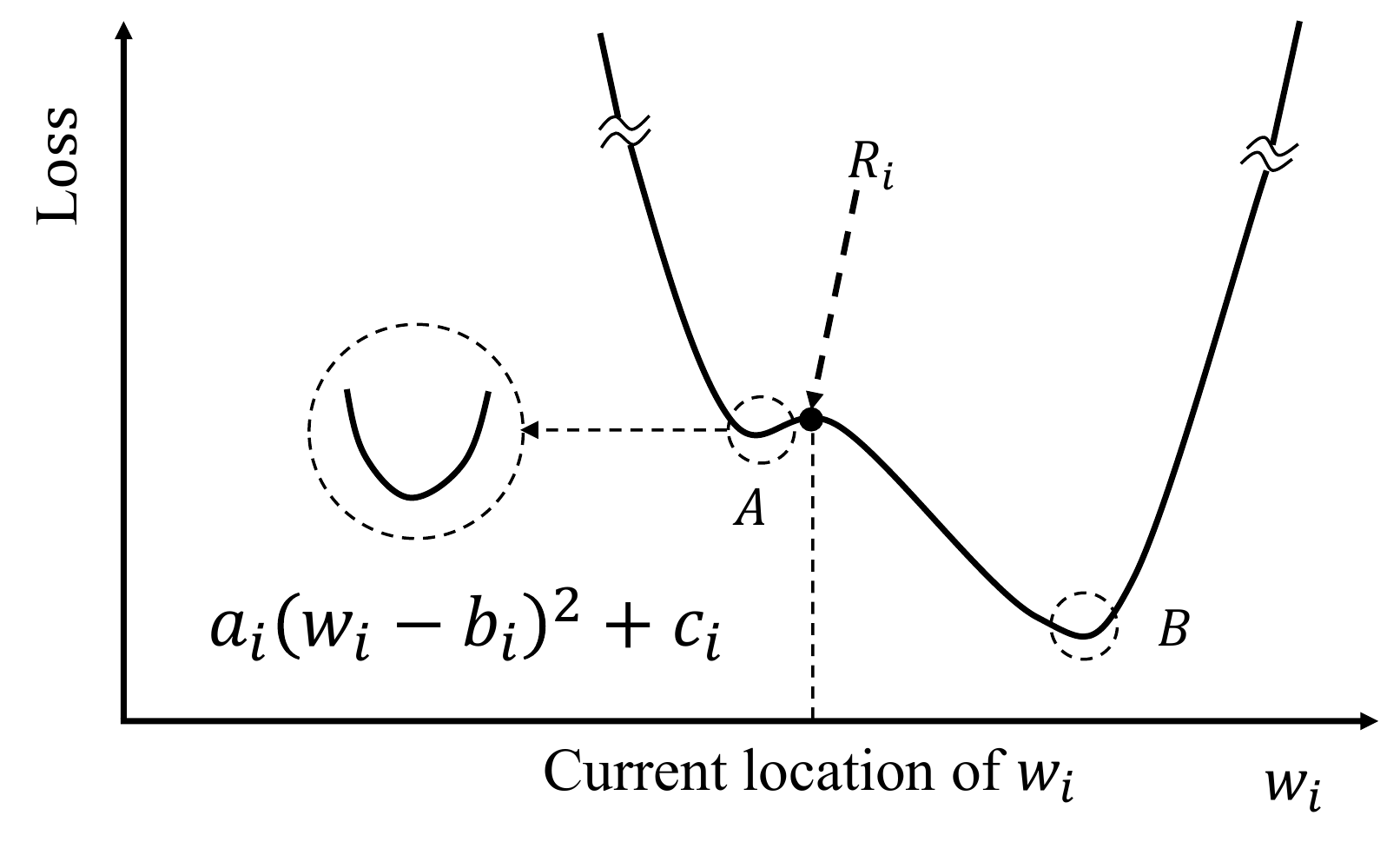}}
% \vskip \figtocaption{}
\vskip -0.1in
\caption{A conceptual sketch of minima. The local geometry equivalents to a parabola.}
\label{fig_15}
\end{center}
\vskip \figrear{}
\end{figure}

Referring to Figure \ref{fig_15}, Morse theory states that in the vicinity of a local critical point of a smooth curve, the geometry is equivalent to a second-order curve, such that

\vskip -15pt %\eqnupspace
\begin{equation}
    \label{eqn_12}
    L=a_i(w_i-b_i)^2+c_i
\end{equation}
\vskip \eqnlowerspace

where \( a_i \), \( b_i \), and \( c_i \) are the characteristic coefficients of the specific parabola.

From this equation we derive the first order gradient:

\vskip \eqnupspace
\begin{equation}
    \label{eqn_13}
    \frac{d L}{d w_i}=2a_i(w_i-b_i)
\end{equation}
\vskip \eqnlowerspace

and second order gradient:

\vskip \eqnupspace
\begin{equation}
    \label{eqn_14}
    \frac{d^2 L}{d w_i^2}=2a_i
\end{equation}
\vskip \eqnlowerspace

Substituting eqn. \ref{eqn_13} into eqn. \ref{eqn_14}, we get

\vskip \eqnupspace
\begin{equation}
    \label{eqn_15}
    \frac{d^2 L}{d w_i^2}=\frac{\frac{d L}{d w_i}}{(w_i-b_i)}
\end{equation}
\vskip \eqnlowerspace

Using this gradient expression, the curvature radius of parameter \( w_i \) becomes

\vskip \eqnupspace
\begin{equation}
    \label{eqn_16}
    R_i = \left| \frac{(w_i-b_i)(1+(\frac{d L}{d w_i})^2)^{\frac{3}{2}}}{\frac{d L}{d w_i}} \right|
\end{equation}
\vskip \eqnlowerspace

where \( w_i \) and \( d L/d w_i \) are detectable quantities, and \( b_i \) is an undetectable coefficients. We propose a simplification here that \( b_i \) can be set to 0, which is reasonable in statistical meanings, especially when weight decay method is used. The \( (d L/d w_i )^2 \) item is also eliminated considering that its value is far less than 1. Then we get

\vskip \eqnupspace
\begin{equation}
    \label{eqn_17}
    R_i\approx \left| \frac{w_i}{\frac{d L}{d w_i}} \right|
\end{equation}
\vskip \eqnlowerspace

This equation is a rough estimate of curvature radius, and it will totally fail in either below cases:

\vskip -15pt %\eqnupspace
\begin{equation}
    \label{eqn_18}
    w_i\rightarrow 0
\end{equation}
\vskip \eqnlowerspace

and

\vskip \eqnupspace
\begin{equation}
    \label{eqn_19}
    \frac{d L}{d w_i}\rightarrow 0
\end{equation}
\vskip \eqnlowerspace

Though the majority of \( R_i \) are not accurate enough for  parameter updating, we estimate some kinds of statistics in a certain group of \( R_i \) are still usable, for example some forms of layer-wise average curvature radius can be used to set the LR.

\subsection{Parameter Optimization Rule Based on the Median of Curvature Radius}

We propose a simple statistical approximation of eqn. \ref{eqn_17} to compute the median value of a group of \( R_i \)

\vskip \eqnupspace
\begin{equation}
    \label{eqn_20}
    R_m\approx \left| \frac{w_m}{\frac{d L}{d w_m}} \right|
\end{equation}
\vskip \eqnlowerspace

where \( w_m \), \( d L/d w_m \) and \( R_m \) are the median values of parameter, gradient and curvature radius in a group (a layer in a NN for example), respectively. Now the LR can be set as

\vskip -15pt %\eqnupspace
\begin{equation}
    \label{eqn_21}
    \eta_g = \gamma R_m
\end{equation}
\vskip \eqnlowerspace

where \( g \) is the group index. If weight decay is considered, instead of eqn. \ref{eqn_20}, we use

\vskip \eqnupspace
\begin{equation}
    \label{eqn_22}
    R_m\approx \left| \frac{w_m}{\frac{d L}{d w_m}+\beta w_m} \right|
\end{equation}
\vskip \eqnlowerspace

where \( \beta \) is the weight decay coefficient.

\begin{figure}[ht]
\vskip \fighead{}
\begin{center}
\centerline{\includegraphics[width=\figwidth]{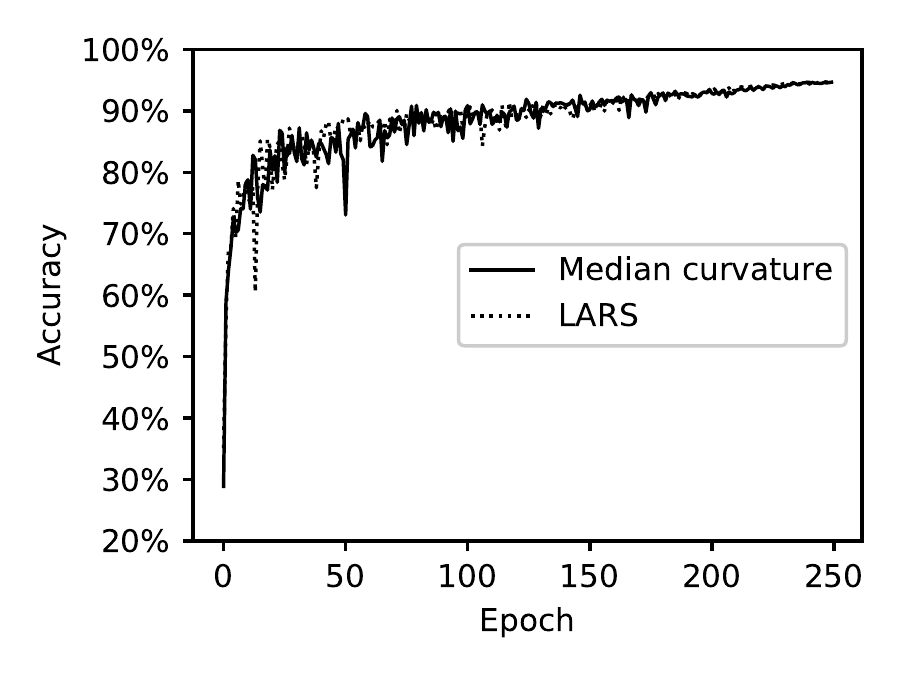}}
\vskip \figtocaption{}
\caption{Training accuracy of median-curvature and LARS with batch-size 1024}
\label{fig_16}
\end{center}
\vskip \figrear{}
\end{figure}

Following the above parameter optimization rule, we have trained ResNet18 on Cifar10 dataset for a series of batch sizes, and find it works well for batch size less than 8192. Figure \ref{fig_16} illustrates the comparation of our method to LARS, and the discrepancy is negligibly small.

We also notice that the training performance falls faster than LARS for large batch size. The reason is the decreasing average gradient (Figure \ref{fig_3}) degrades the validation of \( R_m \) (eqn. \ref{eqn_18}, \ref{eqn_19}, \ref{eqn_20}). Thus, this method can be regarded as a verification of curvature-based parameter update principle, and we do not recommend extending it to large batch size training.

\subsection{Interpretation of Layer-wise Parameter Update Algorithms}

In order to solve large batch size training problem, researchers have proposed one specific class of algorithms, the common point of which is setting a layer-wise LR \cite{you2017large,you2017scaling,abuelhaija2017proportionate,you2018imagenet,you2019reducing}. These algorithms are based on the experimental results, without convincing theoretical foundations. Three typical methods are the so-called LARS \cite{you2017large}, PercentDelt \cite{abuelhaija2017proportionate} and LAMB \cite{you2019reducing}. The definition of the former two algorithms are
LARS,

\vskip \eqnupspace
\begin{equation}
    \label{ean_23}
    \eta_g=\gamma \frac{\|w_i\|_2}{\|\frac{d L}{d w_i}\|_2} 
\end{equation}
\vskip \eqnlowerspace

and PercentDelta,

\vskip \eqnupspace
\begin{equation}
    \label{eqn_24}
    \eta_g=\gamma \frac{size(w_i)}{\| \frac{d L}{d w_i}/w_i \|_1}
\end{equation}
\vskip \eqnlowerspace

where \( size(w_i) \) is the number of parameters in \( g \)th layer. These two algorithms are conjectural solutions to the curvature heterogeneity problem (Figure \ref{fig_2}). Comparing both eqn. \ref{ean_23} and \ref{eqn_24} to eqn. \ref{eqn_17}, it is clear that the main part of LARS and PercentDelta are two transformers of curvature radius (eqn. \ref{eqn_17}). By calculating the norms of the numerator and denominator, two specific statistical mean curvatures are obtained, which can avoid the minimum and maximum problems in the original curvature formula (eqn. \ref{eqn_20}). What the researchers usually ignore is that the algorithms have the same failure conditions, eqn. \ref{eqn_18} and \ref{eqn_19}, which need to be dealt with by careful parameter initialization. 

\section{Discussion}
\subsection{Where Does Parameter Stay in the Loss Curve?}

In this section, we will discuss the geometry meanings of gradient vanishing induced by large batch size training, and will provide a novel insight into how large batch size training degrades the model performance.

Referring to Figure \ref{fig_15}, the distance of \( w_i \) from its nearest minima, \( b_i \), is

\vskip -15pt %\eqnupspace
\begin{equation}
    \label{eqn_25}
    d_i=w_i-b_i
\end{equation}
\vskip \eqnlowerspace

Considering eqn. \ref{eqn_13}, we get

\vskip \eqnupspace
\begin{equation}
    \label{eqn_26}
    d_i=\frac{1}{2a_i} \frac{d L}{d w_i} = \frac{1}{2a_i} g_i
\end{equation}
\vskip \eqnlowerspace

For simplification, we suppose \( a_i \) to be a constant, such that \( a_i=a \). Then, we get

\vskip \eqnupspace
\begin{equation}
    \label{eqn_27}
    d_i\sim N(0,\frac{\sigma}{2a\sqrt{n}})
\end{equation}
\vskip \eqnlowerspace

The expectation of the absolute value of \( d_i \) on all parameters in a layer is

\vskip \eqnupspace
\begin{equation}
    \label{eqn_28}
    E(\left| d_i \right|)=\frac{\sigma}{a\sqrt{\pi}}\frac{1}{\sqrt{n}}
\end{equation}
\vskip \eqnlowerspace

If batch size, \( n \), inflates to an infinity value, this equation predicts that the average distance between a parameter and its nearest minima, \( \left| d_i \right| \), will move to 0. In other words, all the parameters, \( w_i \), in the network will center on a local minimum point of the loss curve, a consequence that makes the parameters optimizing stop and the network training stuck. Making things worse, it happens at the beginning of the training, and the random initialized loss curve strays far from the perfect model that we expect.

One more thing to note is that the vanishing of \( \left| d_i \right| \) also illustrates adopting Morse theory in Subsection 4.1 is valid.

\subsection{How Does Current Methods Work?}

In Introduction section, the main research on ‘generalization gap’ are discussed, and in this section, we try to give a simpler understanding to the main part of these work.

Method 1: LR scheduling \cite{goyal2017accurate,smith2017super-convergence:,peng2017megdet:,you2019large-batch}. This class of methods alleviates the uneven distribution of curvature in different layers (Figure \ref{fig_1}). Taking warm-up training for example, the parameters in a sharp minima (small curvature radius) are given the priority to update, and will find a flatter minima. After warm-up, the curvatures in different layers incline to converge, which is very helpful for the follow-up training.

Method 2: Batch size scheduling \cite{devarakonda2017adabatch:,smith2018don't,Mikami2018imageNet}. Following the conclusion of Subsection 3.2, Figure \ref{fig_4}, and eqn. \ref{eqn_6}, the nature of batch size scheduling is to adjust the parameter update step length, which is an effective way to jump out of sharp minima or noise.

Method 3: Training longer \cite{hoffer2017train}. According to Figure \ref{fig_5} and eqn. \ref{eqn_6}, increasing the iteration steps will enlarge the parameter searching space, which results in a higher probability of finding a better minima.

Method 4: eliminating sharp minima \cite{keskar2017on,jastrzebski2018finding,wen2018smoothout:,haruki2019gradient}. Parameter update step length and curvature of minima are the ‘two faces of the same coin’. For large batch size training, with the vanishing of parameter update step length, big curvature of minima (sharp minima) will play a greater role. Hence, it is evident that eliminating sharp minima will improve the generalization.

\section{Conclusion}

In this work, we have derived the theoretical limits of large batch size training, and predict that ‘generalization gap’ will be worsen with larger batch size. We reveal that the vanishing of gradient and parameter update step length are the key reasons to result in generalization difficulty.

We propose a curvature-based optimizer, a second order LR rule to better fit the curvature variations across layers in a network. We also show that one approximation of this optimizer, median curvature, is almost as good as LARS in small batch size training.

According to our theoretical findings, we suggest a new guidance to reduce ‘generalization gap’ by increasing the gradients in a network. Two algorithms, discarding small-loss samples and scheduling batch size, are provided for example. Our results demonstrate that the generalization is effectively enhanced.

The recent research work on ‘generalization gap’ mentioned in Introduction section have made great experiment advances, but unified theories to rule all the discrete explanations are still inadequate. Our work offers simple theories to support the major parts of current researches. The widely-known LARS and PercentDelta optimizers are proved to be approximations of our curvature-based algorithm. Many other methods are well explained based on the relation between parameter update step length and batch size.

For larger batch size, we find theoretically that the parameters are more likely to center on the bottom of minima, which makes parameter update step length reduced, and finally makes the training to stuck by small minima or noise at a higher probability.

\bibliography{my_paper.bib}
\bibliographystyle{icml2020}

\end{document}